%% file: paper_main.tex
\documentclass[a4paper,fleqn]{cas-dc}
\usepackage[authoryear]{natbib}

\usepackage{float}
\usepackage{multirow}

\usepackage{caption}
\usepackage{subcaption}
\usepackage{capt-of}
\usepackage{multirow}
\usepackage{adjustbox}
\usepackage{amsmath,amsfonts,bm}
\usepackage{alphalph}
\usepackage{amssymb}
\usepackage{pifont}
\usepackage{comment}
\usepackage{amssymb} 
\usepackage{stfloats}
\usepackage{algorithm}
\usepackage{algpseudocode}
\usepackage{xcolor}
\usepackage{pifont}
\usepackage{tabularx}
\usepackage{booktabs}
\usepackage{placeins}
\usepackage{graphicx}
\usepackage{xcolor}
\usepackage{booktabs}

\newcommand{\gcross}{\textcolor{black}{$\boldsymbol{\times}$}}

\def\tsc#1{\csdef{#1}{\textsc{\lowercase{#1}}\xspace}}
\tsc{WGM}
\tsc{QE}
\tsc{EP}
\tsc{PMS}
\tsc{BEC}
\tsc{DE}

\begin{document}
\let\WriteBookmarks\relax
\def\floatpagepagefraction{1}
\def\textpagefraction{.001}

\shorttitle{Land Cover Change in IKN}

\shortauthors{XXX et~al.}

\title [mode = title]{From Forest to Future Capital:
Tracking Land Cover Change in Ibu Kota Nusantara (IKN)  from 2021 to 2026 with PlanetScope Imagery}                      



\author[1]{Clarissa Rui Min Ong}[]
\ead{clarissaong@u.nus.edu}

\author[1]{Elizabeth Tee Inn Loo}[]
\ead{e1355314@u.nus.edu}

\author[1]{Kenneth Woon Hao Soh}[]
\ead{soh.kenneth@u.nus.edu}

\author[1]{William Rachmadi}[]
\ead{williamrachmadi@u.nus.edu}

\author[2]{Qiming Zheng}[type=editor,
                        auid=000,
                        bioid=1,
                        orcid=0000-0002-7393-6585
]
\ead{qmzheng@cuhk.edu.hk}

\author[1]{Hao Li}[type=editor,
                        auid=000,
                        bioid=1,
                        orcid=0000-0002-6336-8772]

\cormark[1]
\ead{Corresponding author: hao.li@nus.edu.sg}

\affiliation[1]{organization={Department of Geography, National University of Singapore},
    city={Singapore},
    postcode={117570}, 
    country={Singapore}
    }

\affiliation[2]{organization={Department of Geography and Resource Management, The Chinese University of Hong Kong},
    city={Shatin, Hong Kong},
    postcode={}, 
    country={China}
    }

\input{0_Abstract}



\maketitle

\input{1_Introduction}

\input{2_Literature_Review}

\input{3_Study_Area_and_Satellite_Data}

\input{4_Methodology}

\input{5_Experiment}

\input{6_Discussion}

\input{7_Conclusion}

\section*{Acknowledgments} This work was supported by the Start-Up Grant (SUG) project “Geospatial Artificial Intelligence for Climate Resilient Urban Environment” from the National University of Singapore (E-109-00-0036-01), and the MoE Tier1 project "Assessing Urban Flood Resilience against Climate Extreme with GeoAI in Southeast Asia".

\section*{Disclosure Statement} No potential conflict of interest was reported by the authors.


\printcredits

\bibliographystyle{cas-model2-names}

\bibliography{interacttfvsample}



\end{document}

%% file: 0_Abstract.tex
\begin{abstract}
     Indonesia's relocation of its political and administrative capital from Jakarta to Ibu Kota Nusantara (IKN) has been framed around a "Forest City" vision, yet rapid development within the Core Government Area (Kawasan Inti Pusat Pemerintahan, KIPP) raises concerns over land conversion, vegetation loss, and carbon stock decline. This study applies remote sensing techniques to systematically assess land conversion and vegetation loss in KIPP from 2021 to 2026 using PlanetScope SuperDove satellite imagery. Full-coverage and cloud-free mosaics were prepared and analysed through spectral indices, including Normalised Difference Vegetation Index (NDVI), Normalised Difference Red Edge (NDRE), and Normalised Difference Water Index (NDWI), alongside supervised Land Use and Land Cover (LULC) classification using a support vector machine algorithm. Results show substantial environmental transformation, with mean NDVI declining by 17.15\%, Total Carbon Stock (TCS) decreasing by 0.28\%, developed land expanding by 670.42\%, and total vegetation declining by 18.14\%. Vegetation loss was most extensive between 2023 and 2024, although a temporary recovery in NDVI and TCS occurred from 2024 to 2025 as active land clearing slowed and development shifted towards already cleared land. Overall, the findings demonstrate that remote sensing provides an effective approach for monitoring the environmental impacts of large-scale urban development, while highlighting the need for higher-resolution, hyperspectral, and synthetic aperture radar-based methods to improve detection of building construction stages and plantation-related land cover changes.
\end{abstract}

%% file: 1_Introduction.tex
\section{Introduction}

For decades, Jakarta has served as the political and economic epicentre of Indonesia. However, rapid urbanisation, severe land subsidence, and chronic environmental degradation have pushed the city beyond its ecological carrying capacity. In response to these mounting pressures, the Indonesian government initiated a historic mega project to relocate the nation's political and administrative centre from Java to East Kalimantan. The new capital, Ibu Kota Nusantara (IKN), is designed not only to alleviate the burden on Jakarta but also to distribute economic development more equitably across the Indonesian archipelago. The relocation represents a monumental shift in spatial planning, transforming large amounts of secondary tropical rainforests, industrial timber plantations, and agricultural land into the administrative heart of the nation \citep{Azizah2024}.

Central to the development of IKN is its ambitious master plan, which brands the new capital as a sustainable, low-carbon "Forest City". Unlike traditional urban redevelopment, the architectural and spatial framework of IKN is predicated on coexisting with nature. The master plan dictates that the capital will encompass approximately 256,000 hectares, with strict land use delineation designed to preserve regional biodiversity \citep{Ardani2024}. In the meantime, the Kunming-Montreal Global Biodiversity Framework establishes urgent 2030 targets to halt ecosystem degradation (Target 1) and enhance urban green spaces (Target 12) \citep{CBD2022gbf}. Herein, the development of IKN involves the conservation and restoration of nearly 200,000 hectares of natural forests, green open spaces, and marine reserves. The built environment is capped at 56,000 hectares of urban areas, strategically interspersed with green corridors. This paradigm aims to maintain at least 65\% to 75\% of the total area as natural tree cover, ensuring that the development minimises its ecological footprint while rehabilitating previously degraded lands in the region.

Despite the sustainable ambition of the IKN master plan, translating these ecological targets into reality requires rigorous, continuous environmental monitoring, especially during the aggressive initial construction phases. The epicentre of this development is the Core Government Area (Kawasan Inti Pusat Pemerintahan, KIPP), which covers 6,671 hectares and has undergone rapid land cover and land use transformation since breaking ground. The aggressive land clearing and infrastructure construction within KIPP raise significant environmental concerns regarding habitat degradation, vegetation loss, and carbon stock decline \citep{Chulafak2024}. Addressing these concerns requires large-scale and high-resolution monitoring beyond theoretical planning frameworks to empirical, spatial accounting of the ongoing physical transformations on the ground.

Though early studies began to document the impacts of this mega project \citep{Aimariyadi2025}, a significant research gap in the current literature is the absence of a time-series, remote sensing (RS)-based quantitative evaluation of IKN's development and its effect on the ecological environment over the development period from early 2020 to date. Existing analyses often focus on isolated environmental indicators, either by evaluating vegetation loss or carbon stock decline separately, or by restricting their temporal scope to the initial months of ground-breaking \citep{Chulafak2024}. As a result, there is a pressing need for an RS-based analytical framework to quantify how the continuous, aggressive construction phases of IKN are altering the region's overall ecological status and carbon storage potential.

Furthermore, monitoring these complex ecological shifts in equatorial forest is severely hampered by persistent atmospheric occlusion and the high spatial heterogeneity of tropical landscapes \citep{Gasparovic2018, Zheng2023}. This methodological challenge exposes a second core research gap: a lack of high-resolution time-series analysis based on openly available RS data and machine learning (ML) techniques \citep{Vizzari2022, Hong2023, Li2024}. Previous evaluations mainly relied on medium-resolution sensors, such as Sentinel-2, which lack the spatial granularity required to detect fine-scale land conversions, narrow infrastructure corridors, and subtle vegetation stress \citep{Li2021, Aimariyadi2025}. Overcoming these bottlenecks requires the integration of high-resolution, high-cadence satellite imagery, featuring the 3-metre PlanetScope SuperDove data, with advanced ML classification algorithms, like the Support Vector Machine (SVM), to accurately classify dynamic land surface transitions \citep{Rodrigues2025}.

To bridge these gaps, this study leverages a continuous time-series of PlanetScope imagery and spectral indices to systematically map land cover dynamics within KIPP from 2021 to 2026 \citep{Chulafak2024}. By integrating high-resolution multispectral data with pixel-based supervised machine learning classification, this research provides an empirical evaluation of how closely actual ground development aligns with IKN's environmental commitments. To this end, this paper aims to address the following three key research questions (RQs):

\begin{enumerate}
    \item How did Land Use and Land Cover (LULC) within KIPP transform during the initial development phase from 2021 to 2026?
    \item What is the quantitative impact of this rapid urban expansion on the tropical forest and its Total Carbon Stock (TCS)?
    \item How effectively can openly available, high-resolution RS time-series and ML capture fine-scale land cover changes in tropical environments?
\end{enumerate}

%% file: 2_Literature_Review.tex
\section{Literature Review}

\subsection{Land Cover Changes via Time-Series Remote Sensing}

Over the past decade, RS-based monitoring of land cover dynamics has fundamentally shifted from static, bi-temporal change detection to dense time-series analysis, largely driven by the opening of satellite archives and the advent of cloud-based geospatial computation platforms like Google Earth Engine \citep{Gorelick2017, Hansen2013}. In tropical regions such as Southeast Asia, persistent cloud cover and atmospheric aerosols have historically restricted the utility of optical remote sensing \citep{Stibig2014}. To mitigate these issues, algorithms capable of capturing continuous land surface phenology and abrupt disturbances, such as the Continuous Change Detection and Classification method, have become standard for medium-resolution data \citep{Zhu2014}.

However, detecting spatiotemporal patterns of forest clearing and rapid urban development requires a spatial granularity for which medium-resolution sensors (e.g., Landsat, Sentinel-2) often prove insufficient \citep{Aimariyadi2025}. As a result, the integration of high-resolution commercial constellations, for example PlanetScope, has emerged as a critical and promising solution. By offering near-daily revisits at a 3-metre resolution, PlanetScope enables the construction of cloud-free mosaics that capture highly heterogeneous land cover changes \citep{Vizzari2022}. Moreover, together with advanced Geospatial Artificial Intelligence (GeoAI) and ML algorithms, such as SVM and Random Forests, these high-dimensional time-series datasets have — albeit only recently — significantly improved classification accuracies for nuanced vegetation mapping, tropical forest monitoring, and rapid disaster response \citep{Rodrigues2025, Li2025}.

Despite these advancements, there is no prior work applying time-series PlanetScope data to monitor land cover changes driven by the development of IKN. Existing assessments have largely been restricted to static, bi-temporal snapshots or very short temporal windows \citep{Aimariyadi2025, Chulafak2024}. As a consequence, there remains a pronounced research gap in deploying ML algorithms over a continuous, multi-year, high-resolution RS time-series. In this context, understanding the feasibility of harnessing openly available, high-resolution commercial RS data becomes a timely research topic. In this paper, we aim to demonstrate how well this framework can resolve fine-scale ecological transitions in complex tropical biomes (RQ3) and accurately chart the trajectory of LULC transformation during KIPP's initial development phase (RQ1).

\subsection{Urban Development and Its Ecological Footprint}

Urban expansion is recognised as one of the most irreversible drivers of global land cover change, heavily impacting regional biodiversity, hydrological cycles, and local climates \citep{Seto2012}. In Southeast Asia, rapid economic growth has catalysed intense urban sprawl, often at the direct expense of highly biodiverse secondary forests and agricultural hinterlands \citep{Estoque2017, Richards2016}. To understand the ecological footprint of this rapid development, researchers frequently look to mature, high-density urban environments like Singapore. Singapore's trajectory from rapid post-independence industrialisation to its current "City in Nature" paradigm highlights both the severe ecological costs of historical habitat fragmentation and the subsequent necessity for aggressive green infrastructure and urban resilience planning \citep{Gaw2019}.

While Singapore provides a longitudinal case study in managing urban ecological limits, mega projects like IKN represent an unprecedented, rapid-scale spatial intervention. The development of IKN within East Kalimantan immediately juxtaposes massive infrastructure demands against sensitive tropical ecosystems \citep{Chulafak2024}. Early RS-based assessments of KIPP's development have documented clear spikes in open land and corresponding vegetation loss, emphasising that even well-intentioned "Forest City" master plans can generate substantial, immediate ecological footprints during their development phases \citep{Aimariyadi2025}.

While the conceptual ecological footprint of such mega projects is widely acknowledged, there is a lack of comprehensive, quantitative evaluation regarding the unique and ongoing effect of IKN development on the local ecological environment over the last five years from 2021 to 2026. Filling this research gap requires transitioning from broad ecological footprint theories to rigorous, pixel-level accounting. By systematically mapping vegetation loss and land cover changes between 2021 and 2026, one can quantitatively assess the immediate environmental impacts of this rapid urban expansion (RQ2).

\subsection{Carbon Stock in Tropical Asia and Its Governance}

Tropical forests are foundational to the global carbon cycle, acting as massive terrestrial carbon sinks; however, their conversion accounts for a significant proportion of global anthropogenic greenhouse gas emissions \citep{Saatchi2011, Baccini2012}. Southeast Asia is particularly vulnerable, as its land use and land cover changes — ranging from logging to the establishment of industrial plantations and urban centres — result in exceptionally high carbon emissions due to the region's dense biomass and carbon-rich peatlands \citep{Ziegler2012}.

Effective and efficient governance of these carbon stocks has become a focal point of regional environmental policy, manifesting in frameworks like Reducing Emissions from Deforestation and Forest Degradation (REDD+) and national nature-based climate solutions \citep{Koh2021}. In evaluating such governance schemes, studies have increasingly scrutinised the alignment between state-led conservation targets and actual ground conditions \citep{Phelps2010}. For IKN, the Indonesian government's commitment to a low-carbon "Forest City" operates as a form of state-led environmental governance. Yet, evaluating the progress of these commitments requires objective, empirical measurement using large-scale time-series RS data.

As highlighted by early analyses, the initial physical construction of IKN has led to measurable declines in local carbon stocks \citep{Chulafak2024}. To fully address this primary research gap, it is important to incorporate continuous high-resolution satellite monitoring with robust carbon estimation models. Existing literature lacks an integrated framework that ties continuous land cover changes directly to both vegetation stress and TCS dynamics over KIPP's multi-year development period. By filling this gap, this paper not only monitors the tangible decline in carbon storage (RQ2) but also establishes an effective, operational framework to audit the ecological promises of IKN's sustainable development master plan.

%% file: 3_Study_Area_and_Satellite_Data.tex
\section{Study Area and Satellite Data}

\subsection{Study Area: Ibu Kota Nusantara}

The study is conducted within the newly designated capital of Indonesia, IKN, located on the east coast of the island of Borneo in the East Kalimantan province. Geographically encompassing portions of the North Penajam Paser and Kutai Kartanegara regencies, the overarching IKN territory spans approximately 256,142 hectares.

The primary spatial focus of this research is restricted to KIPP, as outlined in Figure 1. Covering an area of approximately 6,671 hectares, KIPP serves as the epicentre of the capital's initial construction phase\footnote{For processing purposes, the area of interest used was drawn as a simplified boundary around KIPP, reducing the number of vertices to lower file weight for upload into the Planet Labs web portal. Hence, the classified extent (approximately 8,069 hectares) is larger than the official KIPP boundary.}.
Topographically, the region is characterised by undulating, hilly terrain that was historically dominated by secondary tropical rainforests, industrial timber plantations (primarily \textit{Eucalyptus} and \textit{Acacia}), and agricultural plots.

\begin{figure*}[!ht]
    \centering
    \includegraphics[width=\linewidth]{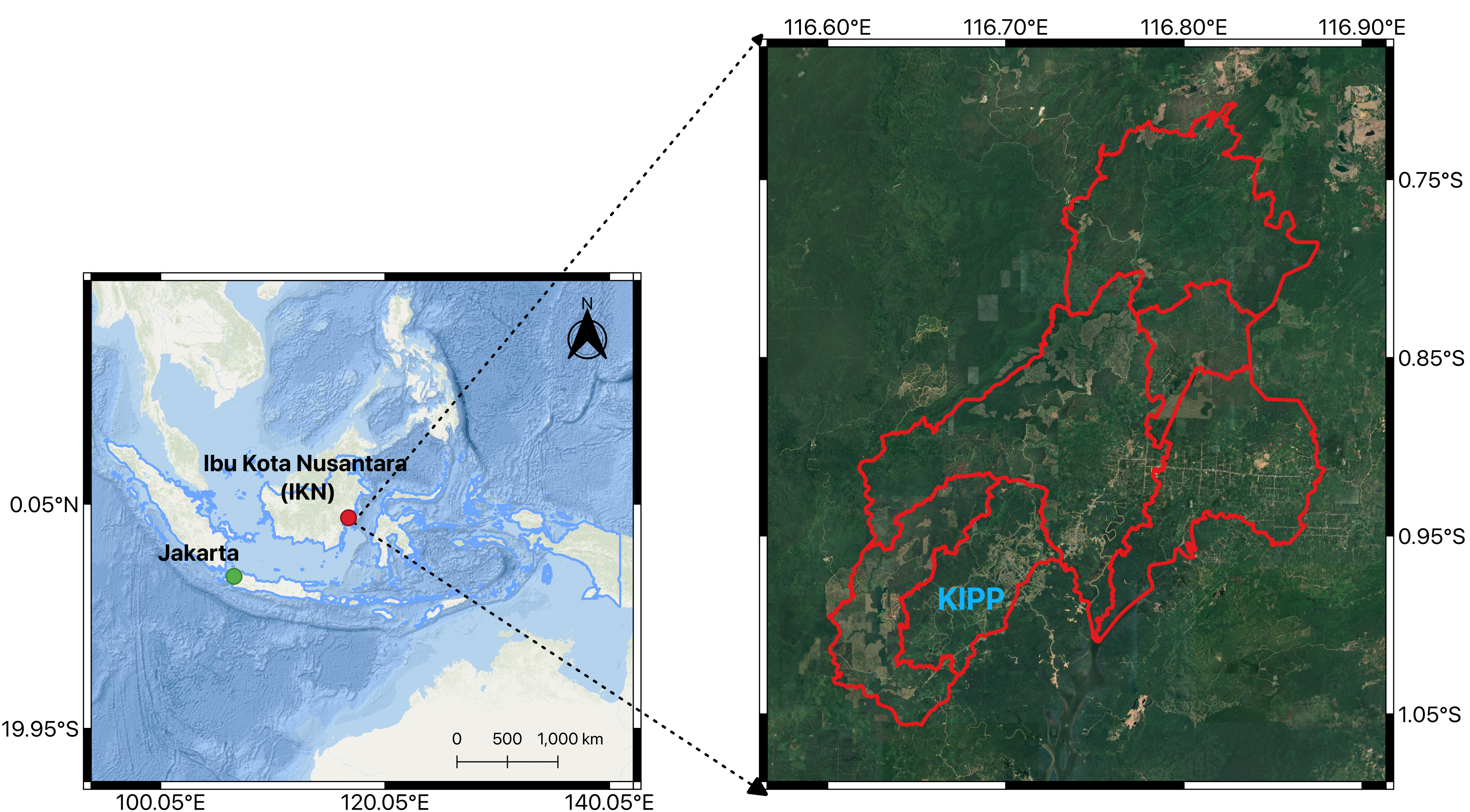}
    \caption{Location of study area. Left: Regional map showing IKN relative to Jakarta. Right: Basemap showing planned boundary of IKN and KIPP.}
\end{figure*}

Monitoring KIPP is critical because it represents the most intense zone of immediate land conversion, specifically housing the Presidential Palace, primary government ministries, and basic infrastructure. Tracking land cover dynamics here provides a localised, high-stakes testing ground for the viability of the overarching ``Forest City'' master plan, which promises to ultimately integrate 56,000 hectares of built urban environment harmoniously within 200,000 hectares of preserved and restored tropical forest.

\subsection{Satellite Data: PlanetScope Multispectral Imagery}

To accurately capture the nuanced land cover transitions resulting from the rapid construction of KIPP, this study utilises high-resolution PlanetScope Multispectral Imagery (MSI) acquired between 2021 (pre-construction baseline) and 2026 (ongoing development phase).

PlanetScope, operated by Planet Labs, consists of a constellation of over 180 CubeSats (Doves and next-generation SuperDoves) deployed in a low Earth, sun-synchronous orbit. The data specifications and advantages for this specific study include:

\begin{itemize}
    \item Spatial Resolution: The imagery provides a ground sampling distance of approximately 3 metres. Unlike medium-resolution data (e.g., 10--30 metres from Sentinel-2 or Landsat), this high spatial granularity is essential for detecting fine-scale land clearings, narrow road networks, and selective logging that characterise the early stages of KIPP's infrastructure development.
    \item Spectral Characteristics: The surface reflectance orthorectified data contains primary bands in the visible, Blue ($0.490\ \mu\mathrm{m}$), Green ($0.560\ \mu\mathrm{m}$), and Red ($0.665\ \mu\mathrm{m}$), as well as the near-infrared spectrum ($0.865\ \mu\mathrm{m}$). The inclusion of the near-infrared band is vital for calculating vegetation indices, such as the Normalised Difference Vegetation Index (NDVI), to continuously monitor forest health, degradation, and vegetation loss.
    \item Temporal Resolution and Cloud Mitigation: East Kalimantan's equatorial climate features persistent and heavy cloud cover, presenting a historical barrier to continuous optical remote sensing. PlanetScope's near-daily global revisit capability serves as a critical advantage. By capturing imagery daily, the probability of acquiring cloud-free observations drastically increases. For this study, the PlanetScope data stream allows for the compilation of continuous, cloud-free yearly mosaics from 2021 to 2026, ensuring that rapid, intra-annual land cover changes are not missed due to atmospheric occlusion.
\end{itemize}

PlanetScope imagery utilised in this study underwent standard radiometric harmonisation and atmospheric correction to bottom of atmosphere surface reflectance to ensure consistency when performing multi-temporal land cover classifications across the five-year observation window.

%% file: 4_Methodology.tex
\section{Methodology}

\subsection{Data Collection}

Satellite imagery used in our study was sourced from Planet Labs' PlanetScope SuperDove constellation, via the Planet Labs Education and Research Program. This fleet consists of multiple flocks of satellites which achieve a near-daily global coverage of MSI at a 3-metre spatial resolution. Using the SuperDove instruments, a 47-megapixel sensor is used to cover 8 spectral bands, namely Red, Green, Blue, Near-Infrared, Green I, Red Edge, Yellow and Coastal Blue. The high spatial, temporal, and spectral resolution of this satellite data allows for detailed monitoring of land use over time.

It should be noted that the Education and Research Program used to access satellite imagery was subject to some constraints. These included a download limit of 3,000 km$^2$ per month, a 30-day publication delay on PlanetScope and RapidEye products, and restricted access to higher-resolution sensors (SkySat, Pelican, and Tanger (hyperspectral)). These limitations did not have a significant impact on our study, as the publication delay does not affect the retrospective analysis conducted in this study, and imagery with a 3-metre resolution was sufficient for the analysis of vegetation and land use in the area.

\subsection{Data Processing}

To prepare clear satellite images for subsequent data analysis, the objective was to obtain a full spatial coverage of the study area with no cloud cover for each year within the study period of 2021 to 2026. This was rarely possible in a single image, due to either incomplete spatial coverage or excessive cloud cover that obstructs the land beneath it. To combat this, a multi-image compositing approach was used. To minimise the effect of seasonality on the images used, all images selected were from the first half of the calendar year, as shown in Table 1 below.

\begin{table}[!htpb]
    \centering
    \begin{tabular}{l c c c c c c c}
        \toprule
        \textbf{Year} & \textbf{Jan} & \textbf{Feb} & \textbf{Mar} & \textbf{Apr} & \textbf{May} & \textbf{Total Imagery} \\
        \midrule
        2021 & & & \gcross & \gcross & & 2 \\
        2022 & & & & \gcross & \gcross & 3 \\
        2023 & & & \gcross & & & 1 \\
        2024 & & & \gcross & & & 1 \\
        2025 & & \gcross & \gcross & & & 4 \\
        2026 & \gcross & & \gcross & & & 3 \\
        \bottomrule
    \end{tabular}
    \caption{Summary of PlanetScope imagery acquisition dates by year.}
\end{table}

By combining the separate images from these dates, a broad-level mosaic was constructed to obtain an image with full coverage and minimal cloud cover. This mosaic was then used as the base layer for further processing. Pixels that were still affected by cloud cover were corrected using ArcGIS Pro's Pixel Editor. For each affected area, clear pixels from an edit layer were identified to replace the original cloud-covered base layer. This process yielded a full-coverage, cloud-free composite that served as the input data for subsequent analysis.

\subsection{Data Analysis}

Following the processing of satellite imagery, spectral composites and indices were derived to examine land surface conditions across our study area. This data was then used for our three main aspects of analysis: NDVI, TCS, and LULC. Figure 2 depicts this process.

\begin{figure*}[!ht]
    \centering
    \includegraphics[width=\linewidth]{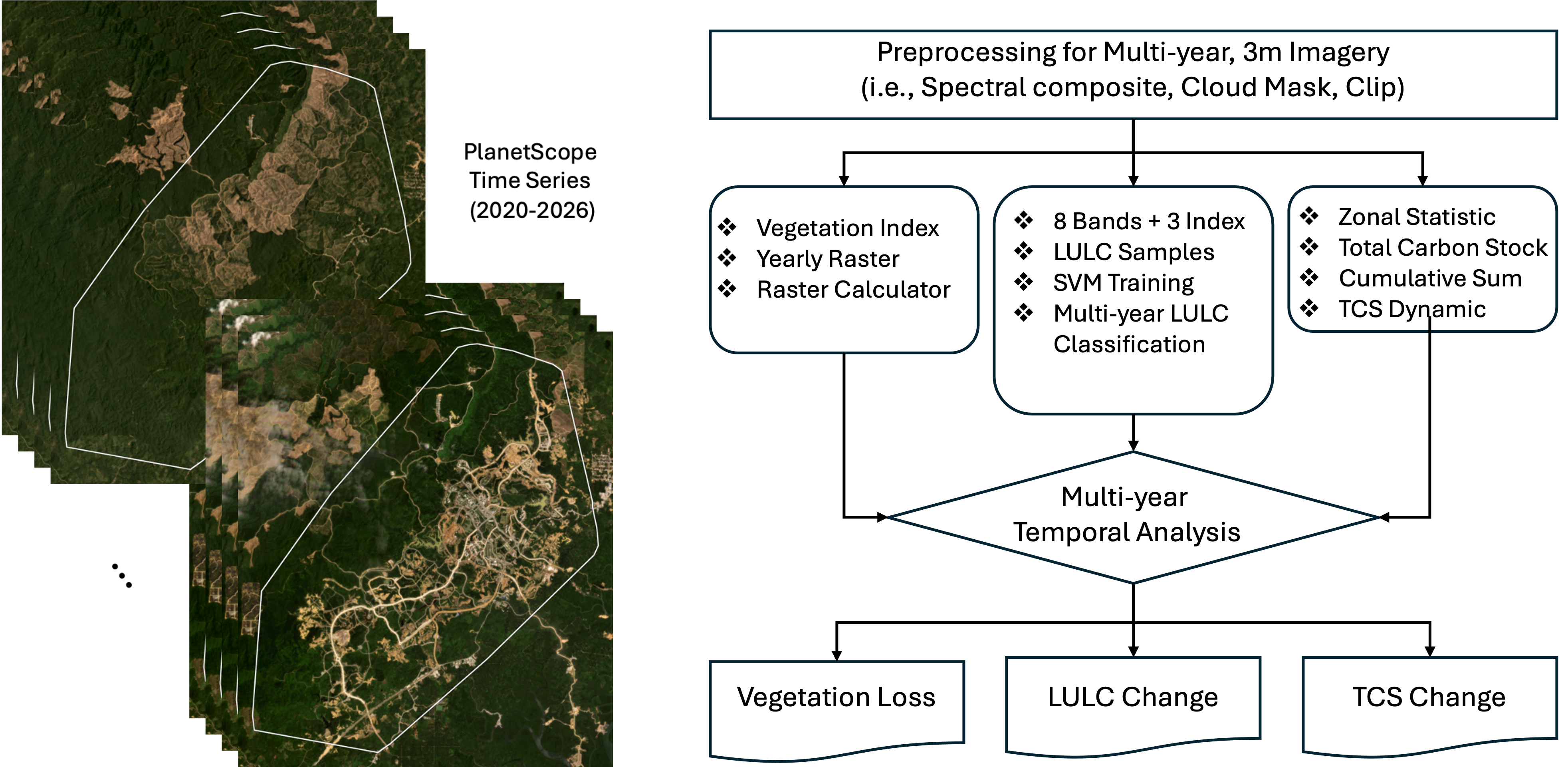}
    \caption{Overall workflow for this study, including PlanetScope imagery processing, LULC classification, together with temporal LULC and TCS analysis.}
\end{figure*}

\subsubsection{PlanetScope Multispectral Imagery Analysis}

For analysis of the satellite data, we used two spectral composites. Natural colour composites (R-G-B) were produced by assigning visible red, visible green, and visible blue bands to their respective display channels, mimicking how human eyes see the world, allowing us to have a visual representation of the area. A false colour composite (NIR-R-G) was also used, where near-infrared, visible red, and visible green bands were assigned to the RGB channels respectively, so as to enhance the spectral contrast between vegetated and non-vegetated areas.

Next, using ArcGIS Pro's Raster Calculator, different spectral indices were calculated across our study area to provide a standardised measure using values ranging from -1 to +1.

\begin{itemize}
    \item \textbf{Normalised Difference Vegetation Index (NDVI)} -- measurement of vegetation condition. By calculating the difference between near-infrared and red bands, NDVI examines vegetation conditions. Higher NDVI values indicate healthier green areas as green vegetation absorbs red light for photosynthesis and strongly reflects near-infrared light. Conversely, negative values can indicate non-vegetated areas or areas with dead vegetation as more red light and less near-infrared light are being reflected. NDVI is calculated as shown in Equation 1.
    \begin{equation}
        \mathrm{NDVI} = \frac{\mathrm{NIR}-R}{\mathrm{NIR}+R}
    \end{equation}
    
    \item \textbf{Normalised Difference Red Edge (NDRE)} -- measurement of vegetation stress. By calculating the difference between near-infrared and red-edge bands, NDRE illustrates vegetation stress through plant chlorophyll content. Higher NDRE values represent plants that have sufficient chlorophyll to absorb red-edge wavelengths while still reflecting near-infrared light, suggesting that they are less stressed. Conversely, lower values indicate stressed or deteriorating vegetation as chlorophyll loss results in less red-edge absorption, increasing red-edge reflectance relative to near-infrared light. This is a useful complement to other indices as it can reflect early-stage vegetation stress before it is visually apparent. NDRE is calculated as shown in Equation 2.
    \begin{equation}
        \mathrm{NDRE} = \frac{\mathrm{NIR}-\mathrm{RE}}{\mathrm{NIR}+\mathrm{RE}}
    \end{equation}

    \item \textbf{Normalised Difference Water Index (NDWI)} -- measurement of water and surface moisture. By calculating the difference between green and near-infrared bands, NDWI shows the presence of water bodies or moisture-saturated areas. Higher NDWI values indicate the presence of such surfaces as water moderately reflects green light while strongly absorbing near-infrared light. Conversely, negative values indicate dry land surfaces, such as vegetation or urban areas, as they reflect more near-infrared than green light. NDWI is calculated as shown in Equation 3.
    \begin{equation}
        \mathrm{NDWI} = \frac{G-\mathrm{NIR}}{G+\mathrm{NIR}}
    \end{equation}
\end{itemize}

The derived spectral composites and indices were then used as a foundation for subsequent analysis.

NDVI analysis acts as a temporally sensitive indicator of vegetation health over our study period of 2021 to 2026, providing a continuous pixel-level record of vegetation gain and loss in terms of both quality and quantity. This process is conducted by assessing the differences between consecutive yearly NDVI data. Using the Raster Calculator to subtract the NDVI value from a given year from that of the following year ($\mathrm{NDVI}_{\mathrm{yr+1}} - \mathrm{NDVI}_{\mathrm{yr}}$), a raster layer for ``Change between years'' is produced. A positive value indicates that there was an increase in vegetation health or density, as more red light is absorbed and more near-infrared light is reflected compared to the previous year. A negative value indicates that the density or condition of vegetation has declined. The results of this analysis showed the differences in the state of vegetation over time through NDVI change maps and descriptive statistics for vegetation in our study area.

\subsubsection{Total Carbon Stock Estimation}

TCS analysis provides a quantitative framework to evaluate the spatiotemporal dynamics of carbon sequestered within terrestrial ecosystems. The terrestrial biosphere plays a pivotal role in the global biogeochemical cycle by acting as a primary carbon sink. In the context of LULC dynamics, transitions such as the conversion of natural vegetation to anthropogenic uses disrupt this equilibrium. As established by \cite{Pan2011}, vegetation loss directly alters terrestrial-atmospheric carbon flux, converting ecosystems from sinks to sources by releasing stored above-ground and soil organic carbon into the atmosphere, thereby exacerbating anthropogenic climate change.

The theoretical basis for utilising vegetation indices as proxies for carbon stock is rooted in the biophysical properties of plant canopies. Foundational remote sensing research demonstrates that the spectral contrast between red band absorption (driven by chlorophyll pigments) and near-infrared reflectance (driven by internal leaf tissue scattering) provides a robust quantification of photosynthetically active biomass \citep{Tucker1979}. This spectral relationship is intrinsically linked to the leaf area index and the fraction of absorbed photosynthetically active radiation, which govern carbon assimilation rates \citep{Sellers1985}. Consequently, NDVI serves as a reliable, spatially continuous surrogate for above-ground biomass and long-term carbon storage capacity.

To operationalise this analysis, carbon stock density is estimated across the spatial domain using an empirically derived exponential model proposed by \cite{Chulafak2024b}. This model establishes a non-linear relationship between NDVI and carbon stock density, capturing the exponential accumulation of biomass in densely vegetated areas. For a given spatial unit (pixel) $i$ at time $t$, the carbon stock density $C_{i,t}$ is computed as:
\begin{equation}
    C_{i,t} = 107.2 \cdot \exp(0.019415 \cdot \mathrm{NDVI}_{i,t})
\end{equation}

To aggregate the TCS across the entire area of interest, spatial integration is performed. The TCS for the region for a given year ($\mathrm{TCS}_t$) is the summation of carbon stock values of all individual pixels:
\begin{equation}
    \mathrm{TCS}_t = \sum_{i=1}^{N} (C_{i,t} \cdot A_i)
\end{equation}
where $N$ represents the total number of pixels within the study area, and $A_i$ denotes the geographical area of pixel $i$.

To monitor the carbon flux resulting from LULC dynamics, the temporal change in TCS ($\Delta \mathrm{TCS}$) between consecutive observation periods is evaluated:
\begin{equation}
    \Delta \mathrm{TCS} = \mathrm{TCS}_{t+1} - \mathrm{TCS}_{t}
\end{equation}

The magnitude and direction of $\Delta \mathrm{TCS}$ serve as robust indicators of ecological state changes and carbon flux. A negative $\Delta \mathrm{TCS}$ ($\Delta \mathrm{TCS} < 0$) indicates a net emission of carbon into the atmosphere (a carbon source), primarily driven by deforestation and land degradation where biomass is depleted. In contrast, a positive $\Delta \mathrm{TCS}$ ($\Delta \mathrm{TCS} > 0$) signifies net carbon sequestration (a carbon sink), indicating an increased capacity of the land to assimilate carbon, typically driven by reforestation, afforestation, or the recovery of natural vegetation.

\subsubsection{Land Use and Land Cover Change Analysis}

The accurate classification of LULC allows for the distinction between various types of land use by assigning land cover categories for each pixel in the study area. Unlike the continuous variables of NDVI and TCS, LULC relies on discrete categorical data to document changes in land use and facilitates identification of areas such as vegetation or cleared land. LULC classification was conducted in three main stages. First, a composite stack was used as the classification input to ensure feature separability, comprising 8 multispectral bands as well as the 3 previously computed spectral indices. Second, we manually selected training samples for five land cover categories: Water, Developed Land, Cleared Land, Sparse Vegetation, and Dense Vegetation. These samples were then used to train a pixel-based supervised SVM classifier using the ArcGIS Pro Image Classification Wizard. This classifier analyses the spectral reflectance of pixels from the given land cover categories, using the training data to draw a hyperplane between classes. Finally, annual differences in land cover were assessed using the ArcGIS Pro Change Detection Wizard, which performed categorical change analysis between outputs for each year, followed by the application of a 3$\times$3 median smoothing filter to reduce salt-and-pepper noise.

%% file: 5_Experiment.tex
\section{Experiment}

In summary, our results reveal three key ecological transformations in KIPP between 2021 and 2026. TCS declined from 976.25 to 973.49 million kg C, a decrease of 0.28\%. Mean NDVI fell from 0.8502 to 0.7044 over the same period, a decrease of 17.15\% (Figure 3(c)). Meanwhile, LULC classification results indicate a decline in vegetated areas alongside an increase in cleared land and developed land (Figure 3(b)).

\begin{figure*}[!ht]
    \centering
    \includegraphics[width=\linewidth]{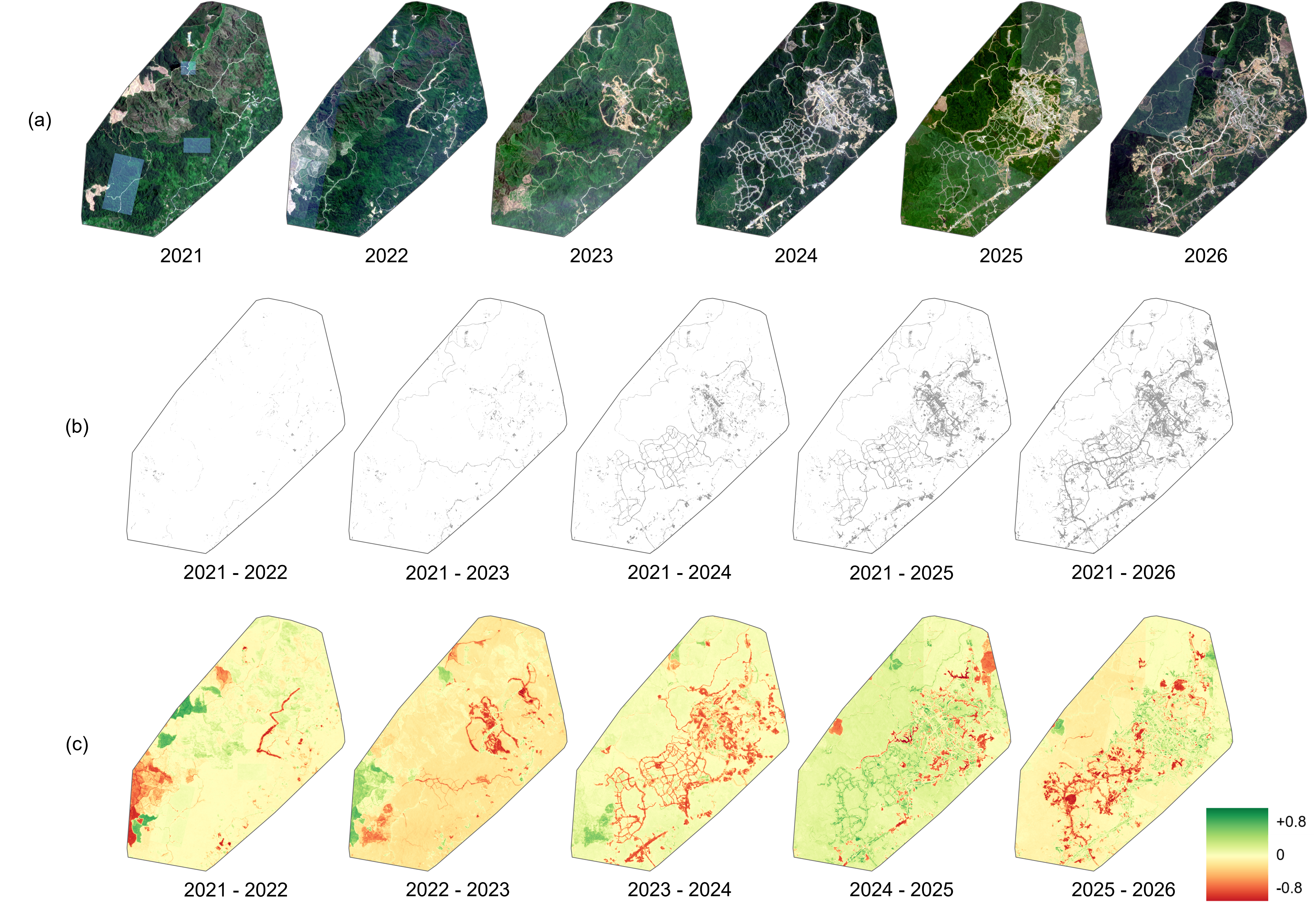}
    \caption{Yearly changes in IKN KIPP (2021–2026): (a) PlanetScope true-colour satellite composites, (b) developed areas changes, and (c) NDVI dynamics.}
\end{figure*}

\begin{table*}[!ht]
\centering
\caption{Annual extent (unit: km$^2$) of LULC classes (2021–2026).}
\begin{tabular}{lrrrrrr}
\toprule
\textbf{Year} & \textbf{Water} & \textbf{Developed Land} & \textbf{Cleared Land} & \textbf{Sparse Vegetation} & \textbf{Dense Vegetation} & \textbf{Total Vegetation} \\
\midrule
2021 & 0.05 & 1.07 & 2.42 & 28.70 & 48.45 & 77.15 \\
2022 & 0.05 & 1.21 & 5.24 & 21.58 & 52.60 & 74.19 \\
2023 & 0.09 & 1.43 & 4.02 & 27.08 & 48.07 & 75.15 \\
2024 & 0.16 & 3.15 & 9.82 & 27.81 & 39.75 & 67.55 \\
2025 & 0.52 & 5.01 & 7.64 & 16.71 & 50.81 & 67.52 \\
2026 & 0.46 & 8.26 & 8.82 & 15.18 & 47.97 & 63.15 \\
\bottomrule
\end{tabular}
\end{table*}

\begin{table*}[!ht]
\centering
\caption{Annual LULC classification accuracy summary (2021–2026).}
\begin{tabular}{lcccccccccccc}
\toprule
\multirow{2}{*}{\textbf{LULC Class}} & \multicolumn{2}{c}{\textbf{2021}} & \multicolumn{2}{c}{\textbf{2022}} & \multicolumn{2}{c}{\textbf{2023}} & \multicolumn{2}{c}{\textbf{2024}} & \multicolumn{2}{c}{\textbf{2025}} & \multicolumn{2}{c}{\textbf{2026}} \\
\cmidrule(lr){2-3} \cmidrule(lr){4-5} \cmidrule(lr){6-7} \cmidrule(lr){8-9} \cmidrule(lr){10-11} \cmidrule(lr){12-13}
 & \textbf{PA} & \textbf{UA} & \textbf{PA} & \textbf{UA} & \textbf{PA} & \textbf{UA} & \textbf{PA} & \textbf{UA} & \textbf{PA} & \textbf{UA} & \textbf{PA} & \textbf{UA} \\
\midrule
Water & 0.98 & 1.00 & 0.96 & 0.98 & 1.00 & 0.68 & 0.96 & 0.86 & 0.94 & 0.98 & 0.94 & 0.98 \\
Developed Land & 0.95 & 0.80 & 0.88 & 0.58 & 0.83 & 0.68 & 0.86 & 0.84 & 0.83 & 0.76 & 0.87 & 0.66 \\
Cleared Land & 0.85 & 0.94 & 0.82 & 0.82 & 0.82 & 1.00 & 0.83 & 0.86 & 0.82 & 0.94 & 0.79 & 0.88 \\
Sparse Vegetation & 0.85 & 0.82 & 0.72 & 0.88 & 0.73 & 0.80 & 0.82 & 0.80 & 0.93 & 0.82 & 0.79 & 0.84 \\
Dense Vegetation & 0.83 & 0.90 & 0.89 & 0.98 & 0.81 & 0.96 & 0.84 & 0.92 & 0.94 & 0.96 & 0.92 & 0.94 \\
\midrule
\textbf{Overall Accuracy} & \multicolumn{2}{c}{\textbf{0.892}} & \multicolumn{2}{c}{\textbf{0.848}} & \multicolumn{2}{c}{\textbf{0.824}} & \multicolumn{2}{c}{\textbf{0.856}} & \multicolumn{2}{c}{\textbf{0.892}} & \multicolumn{2}{c}{\textbf{0.860}} \\
\textbf{Kappa Coefficient} & \multicolumn{2}{c}{\textbf{0.865}} & \multicolumn{2}{c}{\textbf{0.810}} & \multicolumn{2}{c}{\textbf{0.780}} & \multicolumn{2}{c}{\textbf{0.820}} & \multicolumn{2}{c}{\textbf{0.865}} & \multicolumn{2}{c}{\textbf{0.825}} \\
\bottomrule
\multicolumn{13}{l}{\small \textit{Note:} PA = Producer Accuracy; UA = User Accuracy.} \\
\end{tabular}
\end{table*}

\subsection{Land Use and Land Cover Classification Accuracy}

Table 2 and Table 3 report the LULC areas and classification accuracy performances for each year from 2021 to 2026, respectively. Based on four classic accuracy metrics, namely user's accuracy, producer's accuracy, overall accuracy, and the Kappa coefficient, we have the following observations: First, overall accuracy ranges from a minimum of 82.40\% to a maximum of 89.20\%, with a mean of 86.20\% across the study period, which confirms the capability of PlanetScope MSI in the LULC classification task. Secondly, the Kappa coefficient similarly ranges from a minimum of 0.78 to a maximum of 0.87, with a mean of 0.83, showing that the input feature space and the LULC classes achieve a balanced result.

Moreover, as each yearly confusion matrix is built on an equal sample of fifty ground-truth points per class, mean user's accuracy and mean producer's accuracy converge closely with the overall accuracy figure in every year of the study period. This indicates that classification error is distributed relatively evenly across classes, rather than concentrated in the omission or commission of any single land cover type, and suggests that no individual class systematically undermines the reliability of the classification as a whole.

These results compare favourably with recent benchmarking of comparable classifiers; for instance, \cite{Zafaretal2024}, in an assessment of machine learning algorithms for LULC mapping using Google Earth Engine, report a SVM overall accuracy of 89.80\% and a Kappa coefficient of 0.84 as the ceiling achieved by that classifier across their study period. Our SVM classification, with a mean overall accuracy of 86.20\% and mean Kappa coefficient of 0.83, falls within a comparable range to this benchmark.

To sum up, the LULC classification and change detection results indicate that our method is sufficiently accurate and consistent to support the LULC trends reported in this study. Despite the methodological limitations in our discussion portion later on, namely the patchwork effect introduced by mosaic-and-edit compositing, the confounding influence of plantation cycles, and the undetected development occurring within the developed land class, the stability of accuracy across six independent yearly classifications indicates that these limitations constrain the precision of specific transitions rather than undermining the overall reliability of our results.

\subsection{Land Use and Land Cover Change and Total Carbon Stock Dynamics}

Table 4 shows the results of IKN's LULC change dynamics between 2021 and 2026, and reveals significant spatiotemporal transformations of LULC within the KIPP region, as shown in Figure 4. Monitoring these changes provides quantitative insight into the longitudinal environmental impacts of the rapid construction phase associated with IKN's development. Moreover, the observed LULC change trend demonstrates a clear trajectory of continuous human-environment intervention, featured primarily by the aggressive expansion of built environments and the corresponding depletion of natural land cover. Two key findings deserve special attention herein.

The first key finding highlights an exponential expansion of urban infrastructure and land clearing. By 2026, developed land (as shown in Figure 3(b)) experienced a massive increase of 670.42\%, equating to an absolute expansion of 7.18 km\textsuperscript{2} relative to the 2021 baseline. In addition, cleared land expanded by 264.57\%, or 6.40 km\textsuperscript{2}, confirming the aggressive pace of land conversion required to support the early stages of the mega project.

From an urban growth perspective, the second key finding reveals a compounding and consistent loss of regional vegetation. Total vegetation cover declined consecutively each year, resulting in an absolute reduction of 14.00 km\textsuperscript{2} (i.e., a decrease of 18.14\%) by 2026. As shown in Figure 5(a), this ecological shift was most heavily driven by the degradation of sparse vegetation, which saw a severe 47.11\% drop (i.e., -13.52 km\textsuperscript{2}), reflecting the immediate ecological impact of the ongoing infrastructure developments.

\begin{figure*}[!ht]
    \centering
    \includegraphics[width=\linewidth]{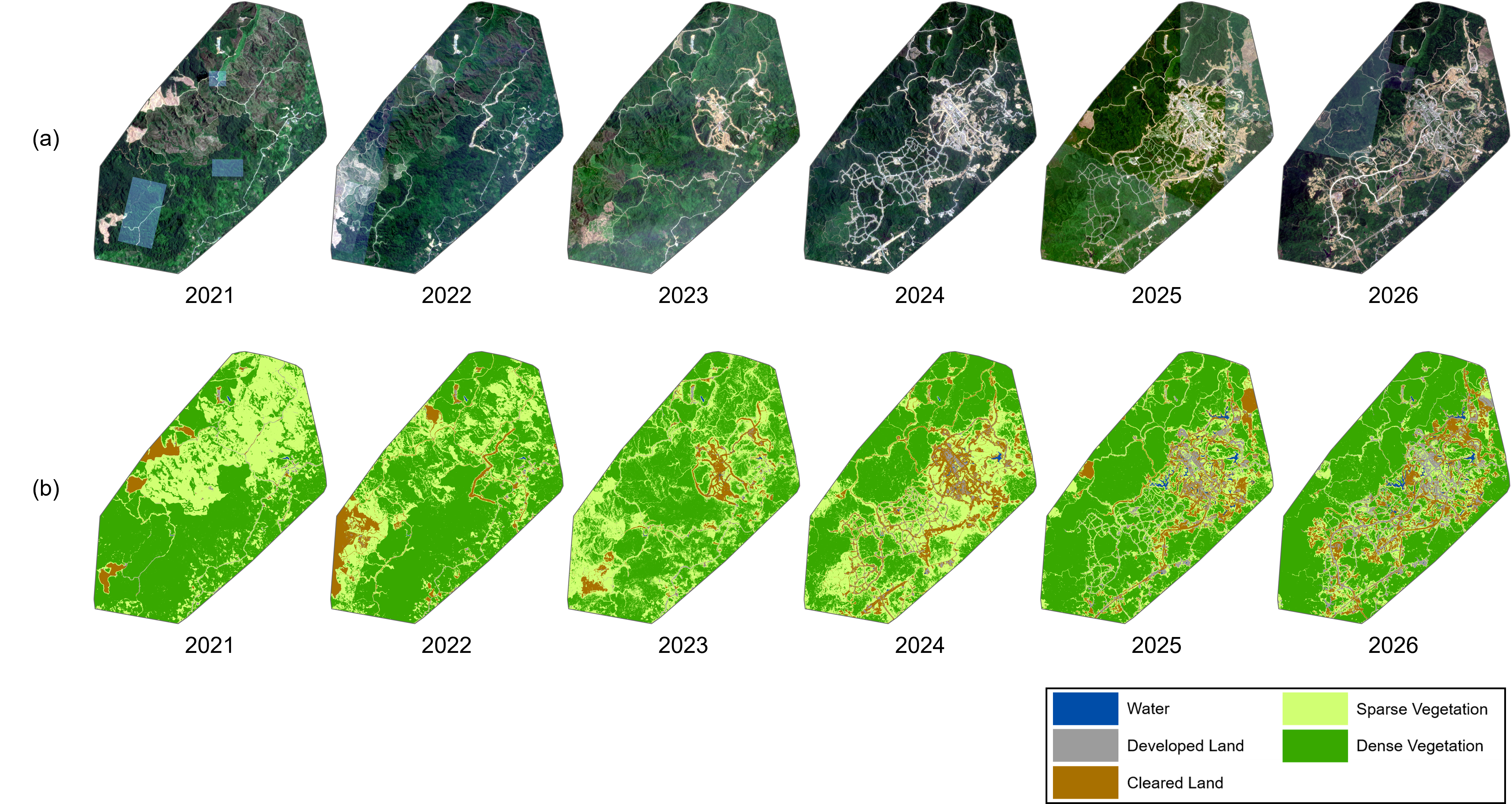}
    \caption{Yearly LULC Classification in IKN KIPP (2021–2026): (a) PlanetScope true-colour satellite composites and (b) LULC classification maps.}
\end{figure*}

\begin{table*}[!ht]
\centering
\caption{LULC change dynamics relative to the 2021 baseline.}
\begin{tabular}{lrrrrrr}
\toprule
\textbf{Year} & \textbf{Water} & \textbf{Developed Land} & \textbf{Cleared Land} & \textbf{Sparse Vegetation} & \textbf{Dense Vegetation} & \textbf{Total Vegetation} \\
\midrule
\multicolumn{7}{l}{\textit{Absolute Change (km\textsuperscript{2})}} \\
\midrule
2022 & $-0.00$ & $+0.14$ & $+2.82$ & $-7.12$ & $+4.16$ & $-2.96$ \\
2023 & $+0.04$ & $+0.36$ & $+1.60$ & $-1.62$ & $-0.38$ & $-2.00$ \\
2024 & $+0.11$ & $+2.08$ & $+7.40$ & $-0.89$ & $-8.70$ & $-9.59$ \\
2025 & $+0.47$ & $+3.94$ & $+5.22$ & $-11.99$ & $+2.37$ & $-9.63$ \\
2026 & $+0.41$ & $+7.18$ & $+6.40$ & $-13.52$ & $-0.48$ & $-14.00$ \\
\midrule
\multicolumn{7}{l}{\textit{Percentage Change (\%)}} \\
\midrule
2022 & $-6.12$ & $+13.19$ & $+116.62$ & $-24.81$ & $+8.58$ & $-3.84$ \\
2023 & $+74.69$ & $+33.78$ & $+66.04$ & $-5.65$ & $-0.77$ & $-2.59$ \\
2024 & $+229.80$ & $+194.01$ & $+305.75$ & $-3.11$ & $-17.96$ & $-12.43$ \\
2025 & $+966.53$ & $+367.37$ & $+215.47$ & $-41.79$ & $+4.89$ & $-12.48$ \\
2026 & $+834.29$ & $+670.42$ & $+264.57$ & $-47.11$ & $-0.98$ & $-18.14$ \\
\bottomrule
\end{tabular}
\end{table*}

To confirm these findings, Figure 5 shows the relative temporal dynamics of TCS along with the changes of different LULC classes. Herein, Figure 5(a) demonstrates a distinct expansion in anthropogenic land uses, characterised by the steady upward trajectories of both developed land and cleared land. In contrast, natural land covers exhibit a marked decline. In the meantime, dense vegetation and sparse vegetation show oscillating patterns that culminate in a substantial net decrease in total vegetation. Meanwhile, the relative extent of water remains largely invariant, maintaining a stable baseline from 2021 to 2026.

Corresponding to LULC changes, Figure 5(b) delineates the cumulative impact on IKN's TCS. The temporal curve reveals an overall downward trend in carbon stocks, reflecting the loss of vegetative cover and the concurrent land clearing processes. Although a transient period of partial recovery is observable in the latter half of the time series, the overall dynamic features a significant decrease in carbon storage capacity. This inverse correlation emphasises the significant environmental impact of replacing natural vegetation with developed or cleared surfaces.

\begin{figure*}[!ht]
    \centering
    \includegraphics[width=\linewidth]{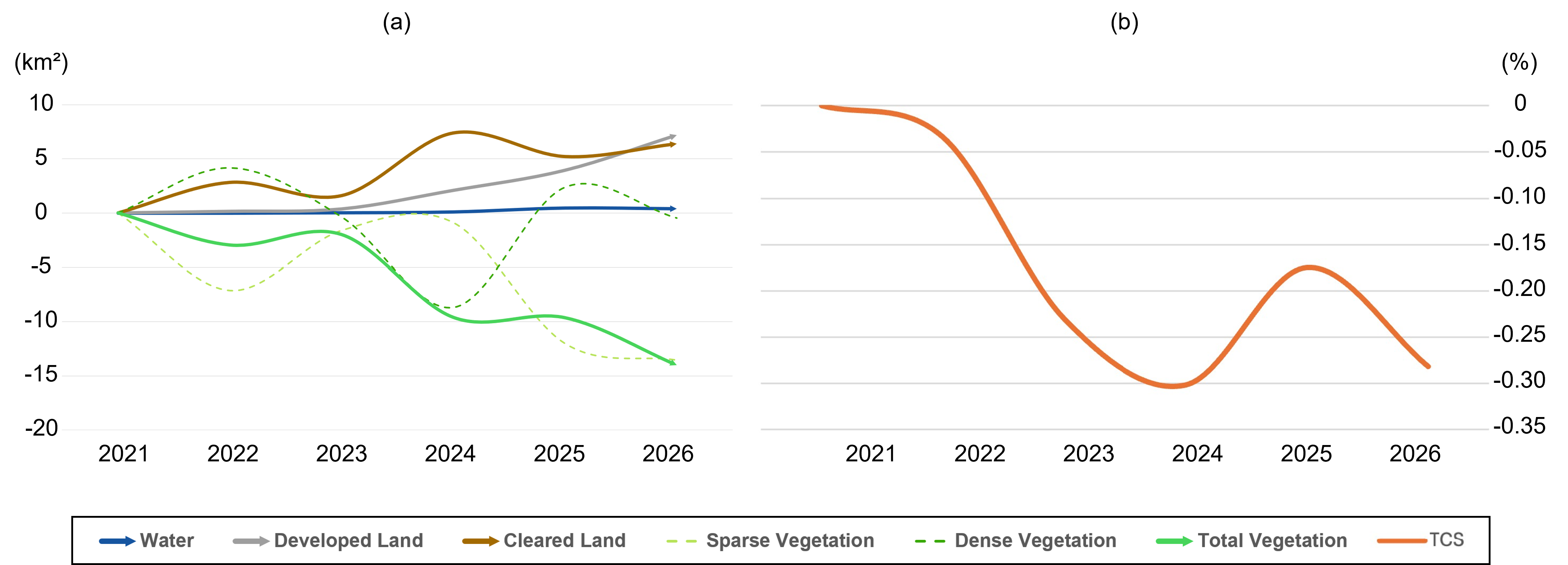}
    \caption{Temporal dynamics of (a) LULC changes and (b) TCS changes from 2021 to 2026.}
\end{figure*}

\subsection{Applicability to Local Stakeholders}

Beyond these experimental results, this study showcases a practical workflow of time-series remote sensing that extends to environmental organisations, investors, and consultants seeking to monitor the development of a new capital city independently of official reporting.

For environmental organisations, this study offers a means of independently verifying progress against IKN's own environmental commitments. The Nusantara Capital Authority and Ministry of National Development Planning, in the government official master plan, commit the project to a ``Smart, Green, Beautiful, and Sustainable City'' vision, delivered through a ``Forest City'' concept \citep{PressRelease2020}. The NDVI, TCS, and LULC methodology developed here provides environmental organisations with a practical tool for tracking vegetation loss and regrowth against this stated vision over time, rather than relying solely on officially reported figures.

For investors and consultants, this study addresses a related but distinct need: assessing the credibility and pace of a multi-decade mega project to inform decisions on whether to commit capital or prepare to shift business functions to the region. This is a documented dilemma rather than a hypothetical one. Edelman Global Advisory \cite{Edelman2023}, in an analysis of the investment climate surrounding IKN, notes that investor confidence depends on ``clearly identified and well-orchestrated information sources'' for delivering progress updates, and that a stable investment climate is essential to catalyse investors' decision-making process in participating in large-scale projects. The developed-land and vegetation-loss trends reported in this study offer precisely this kind of independent, verifiable progress signal, and could help inform investors and consultants navigating this decision.

%% file: 6_Discussion.tex
\section{Discussion}

In this paper, we systematically examined land cover transformations in IKN's Core Government Area from 2021 to 2026. To address our research questions, we demonstrated that integrating PlanetScope time-series MSI with ML methods effectively captured fine-scale tropical LULC changes. Our analysis quantified the immediate environmental impact of the capital's construction, revealing a 670.42\% expansion in developed land, an 18.14\% decline in total vegetation cover, and a 0.28\% decline in TCS. Despite these important findings establishing a robust baseline for continuous monitoring, we also identified several methodological limitations to be addressed in future work.

\subsection{Obtaining Full-Coverage, Cloud-Free Imagery}

The persistent presence of cloud cover, cloud shadow, and atmospheric haze in East Kalimantan presented a significant challenge to our data processing. Obtaining PlanetScope imagery with complete spatial coverage of KIPP and minimal atmospheric interference for a given year proved difficult using any single acquisition. To address this, each yearly composite, except for 2023 and 2024, for which a single full-coverage, cloud-free image was available, was constructed using a broad-level mosaic-and-edit method. A base image acquired between January and June was selected on the basis of maximum spatial coverage and minimal atmospheric interference, with remaining coverage gaps resolved by mosaicking additional scenes within Planet's platform, and residual cloud, shadow, or haze corrected using ArcGIS Pro's Pixel Editor, drawing on clearer alternatives from other acquisition dates.

Despite median compositing being widely regarded as the more rigorous alternative method for obtaining full-coverage, cloud-free imagery, it produced less-than-satisfactory results in this study when tested as an alternative to the mosaic-and-edit approach. For each pixel, this method selects the median value across all available observations for a given period, with the intention that transient artefacts such as cloud, shadow, and haze would be statistically excluded in favour of the underlying land surface reflectance. In East Kalimantan, however, the prevalence of atmospheric interference across the available image stack meant that a substantial proportion of candidate observations were themselves contaminated, such that the resulting median value at a given pixel frequently corresponded to a contaminated rather than a clear observation. This produced composites of visibly lower quality than those built through mosaicking, retaining substantial atmospheric and shadow artefacts across large parts of the study area. Mean compositing was also tested as a secondary alternative, but proved similarly unsuitable, as averaging pixel values across contaminated and clear observations produced spectrally mixed values that did not correspond to any real land cover class.

We further considered pre-masking cloud, shadow, and haze pixels prior to compositing, which in principle would allow statistical methods such as median compositing to perform more reliably. Planet's Usable Data Mask 2 (UDM2) was evaluated for this purpose but was found to inconsistently flag affected pixels, particularly thin haze and shadow. Manually delineating masks for each scene across the full study period was determined to be beyond the scope of this study, given the number of images involved.

The mosaic-and-edit method proved more effective for this study because it does not rely on the statistical prevalence of clear observations within an image stack. Each region of the composite is instead drawn from an individually verified, unambiguously clear scene, rather than an aggregated value computed across a contaminated stack. On this basis, we retained the mosaic-and-edit method as our primary data processing approach, rather than the median or mean compositing methods tested. This yields a full-coverage, cloud-free composite for each year, at the cost of a discernible patchwork effect where the composite draws on imagery acquired on different dates.

\subsection{Confounding Industrial Plantation Cycles}

Industrial plantations within the study area presented a confounding factor throughout this study. Plantation land cover follows cycles of clearing and regrowth that operate independently of urban development activities, and these cycles are registered within our classification as fluctuations in cleared land, sparse vegetation, and dense vegetation. This affects the specific NDVI, TCS, and LULC values reported for these three classes, and means that the change observed between 2021 and 2026 cannot be attributed to development activity alone.

This limitation does not, however, undermine the overall validity of our results. Plantation-driven fluctuations introduce noise around the margins of the three classes rather than displacing the dominant signal identified in this study, namely the sustained loss of overall vegetation and corresponding gain in developed land.

A more precise disaggregation of these two drivers would nonetheless be possible with hyperspectral imagery, which captures a substantially greater number of narrow, contiguous spectral bands than the multispectral sensor employed here. This would enable material-level discrimination between plantation-driven change, such as the clearing of mature plantation stands or the sparse early regrowth of replanted areas, and development-driven vegetation disturbance, such as the degradation of vegetation at the margins of newly cleared land or the horticultural regrowth planted around completed development areas. Incorporating hyperspectral data in future work would allow these plantation-driven changes to be disaggregated from development-driven change directly, rather than treated as an acknowledged source of confounding noise.

\subsection{Undetected Building Development}

Treating developed land as a single classification class limits detection of intermediate construction activity within that class. A structure that increased from one to five storeys between two years, for instance, would be classified as developed land in both years, such that consecutive-year results show no apparent change despite substantial construction activity having occurred. This limitation risks understating the pace of IKN's development in later stages, when growth increasingly takes the form of vertical construction rather than new land clearing and development.

Notwithstanding this limitation, the overall trajectory of IKN's development remains clearly captured by our classification, as the transition from vegetation and cleared land to developed land observed across the 2021 to 2026 period constitutes the dominant and unambiguous signal in our results.

Two approaches could address this finer-grained limitation in future work. The first pairs higher-spatial-resolution imagery with a classification algorithm trained to distinguish discrete construction stages, rather than a binary developed or undeveloped classification. The second is Synthetic Aperture Radar (SAR), an active sensor that measures the physical geometry of built structures directly, rather than their surface reflectance, thereby enabling structural change to be detected independent of cloud cover or lighting conditions. Incorporating either approach would allow construction progress at IKN to be tracked with considerably greater granularity than our current classification permits.

%% file: 7_Conclusion.tex
\section{Conclusion}

In this paper, we demonstrate the critical advantages of PlanetScope MSI in monitoring IKN's rapid urban development from 2021 to 2026. Specifically, the 3-metre spatial resolution proved effective in detecting fine-scale LULC changes, such as narrow infrastructure corridors and early-stage land clearing, which are often missed by medium-resolution RS imagery. By integrating PlanetScope MSI data with SVM algorithms and multispectral indices (e.g., NDVI, NDRE, NDWI), we achieved a robust 86.20\% mean overall classification accuracy over five LULC classes.

Moreover, the temporal analysis of LULC changes alongside TCS dynamics reveals the immediate ecological impact of IKN's development. From 2021 to 2026, developed land increased by 670.42\%, driving an 18.14\% decrease in total vegetation cover. This rapid urbanisation reduced TCS by 0.28\%, from 976.25 to 973.49 million kg C. Ultimately, this PlanetScope-based monitoring workflow provides stakeholders with a timely and reliable solution to verify the government's "Forest City" commitments. Future monitoring could consider integrating higher-resolution, hyperspectral, and SAR imagery to better distinguish LULC changes caused by cyclical plantation harvesting and detect different stages of building construction.